\documentclass[sigconf,authorversion]{acmart}
\usepackage{amsmath}

\AtBeginDocument{%
  }

\setcopyright{rightsretained}
\copyrightyear{2024}
\acmYear{2024}
\acmConference{SA Posters '24}{December 03-06, 2024}{Tokyo, Japan}
\acmBooktitle{SIGGRAPH Asia 2024 Posters (SA Posters '24), December 03-06, 2024}
\acmDOI{10.1145/3681756.3697973}
\acmISBN{979-8-4007-1138-1/24/12}
\begin{document}

\title{Neural Clustering for Prefractured Mesh Generation in Real-time Object Destruction}

\author{Seunghwan Kim}
\affiliation{%
  \institution{Kyung Hee University}
  \city{Yongin}
  \country{South Korea}}
\email{overnap@khu.ac.kr}

\author{Sunha Park}
\affiliation{%
  \institution{Kyung Hee University}
  \city{Yongin}
  \country{South Korea}}
\email{icecream20@khu.ac.kr}

\author{Seungkyu Lee}
\affiliation{%
  \institution{Kyung Hee University}
  \city{Yongin}
  \country{South Korea}}
\email{seungkyu@khu.ac.kr}


\begin{abstract}
    Prefracture method is a practical implementation for real-time object destruction that is hardly achievable within performance constraints, but can produce unrealistic results due to its heuristic nature. To mitigate it, we approach the clustering of prefractured mesh generation as an unordered segmentation on point cloud data, and propose leveraging the deep neural network trained on a physics-based dataset. Our novel paradigm successfully predicts the structural weakness of object that have been limited, exhibiting ready-to-use results with remarkable quality.
\end{abstract}



\begin{teaserfigure}
  \includegraphics[width=\textwidth]{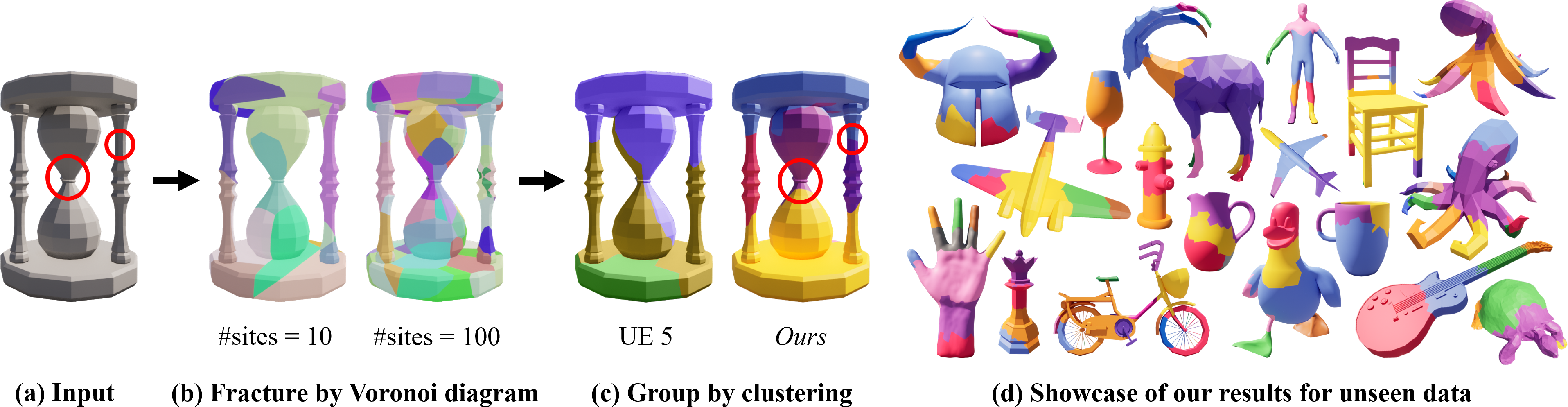}
  \caption{Prefracuterd mesh generation and results. Each piece and group is colored randomly. (a) Input mesh with highlighted structural weakness, thin points marked in red circles. (b) Prefractured meshes made solely from Voronoi diagrams, according to the number of sites. (c) Prefractured meshes refined from (b) with auto-clustering in Unreal Engine 5 (UE 5) and our method. Only ours reflects structural weaknesses at the designated positions, shown in red circles, whereas Voronoi-only and UE 5 do not. (d) Results of ours for unseen data from Objaverse \cite{deitke2023objaverse}. Each prefractured mesh considers the structure of object, for example assigning fractures to extremities such as wings, horns, and legs, which are prone to breakage.}
  \Description{Prefractured meshes. (a) Input mesh that is the shape of an hourglass. (b) Randomly prefractured mesh. (c) Refined prefractured mesh. (d) A lot of prefractured meshes made by the proposed method.}
  \label{fig:teaser}
\end{teaserfigure}
% \received{11 August 2024}
% \received[revised]{7 October 2024}

\maketitle

\section{Introduction}

% Prefracutre의 일반적인 과정 (간략하게!)
Real-time object destruction is challenging due to computational complexity. Fully physics-based simulations remain expensive despite recent efforts \cite{huang2023deepfracture}. \textit{Prefracture method} is a feasible implementation of real-time object destruction, which is adopted in practical applications like Unreal Engine 5 (UE 5). It eludes simulation at runtime by utilizing a pre-computed fractured version of the original mesh. When an impact is detected at runtime, the mesh is entirely or partially replaced with its prefractured counterpart. The remainder is rigidbody simulation with its fragments, incurring reasonable cost.

% 현재 상용 기술(언리얼 엔진)의 한계, 개선가능성
However, it relies on heuristics and thus can cause unrealistic destruction, as depicted in Figure \ref{fig:teaser}. The generation of prefracutred mesh typically involves breaking the mesh into fine pieces using Voronoi diagram, and then clustering them into larger pieces that are called \textit{group} in this work. Various heuristics are used in this process for aesthetic results, but they often fail to reflect the structural weakness of object \cite{sellan2023breakinggood}. Currently, this is manually corrected by an artist, which is very labor-intensive. This work addresses the issue using a deep neural network that learns from a physics-based object destruction dataset.

\section{Method}

% Point Transformer를 사용함 (간단하게 구조 설명하고 사용한 이유)
We introduce the network into the clustering stage in prefractured mesh generation. Since it is difficult to handle the mesh directly, we gather the center of mass from pieces to construct point cloud. Our goal is clustering point cloud with ground truth group labels from datasets, which can be generalized to point cloud segmentation. We employ \textit{Point Transformer} \cite{zhao2021point} to predict a group label, as indicated in Figure \ref{fig:architecture}. It provides U-net architecture with efficient self-attention on point cloud and shows competitive performance, leading to becoming a baseline for recent point cloud research. We use the default hyperparameter as proposed in the work, except for minor modifications to fit the shape of in/output. % 디테일 추가?

% Unordered label 때문에 그냥은 못씀
However, the group label is distinct from the conventional segmentation label. We only focus on the identity between each point rather than the numerical value of group label. Namely, the group label is \textit{unordered}. Hence the loss must be permutation-invariant in both the prediction and ground truth group label; i.e., remaining consistent if labels share the same group structure across points.

% Unordered label을 Probabilistic modeling으로 해결
To resolve the unordered label, we design the loss based on pairwise identity, assuming point-to-point independence. For any points $x_i, x_j \in X$, we define the softmax of network output as the probability that each point belongs to a particular group $k \in G$, $p(g(x_i)=k)$, where $g(x_i)$ denotes the group label of $x_i$ in an arbitrarily determined order. Assuming $p(g(x_i)=k)$ and $p(g(x_j)=k)$ are independent, $p(g(x_i)=g(x_j))=\sum^G_k{p(g(x_i)=k)p(g(x_j)=k)}$ holds. From the ground truth group label in an arbitrarily determined order $y_i, y_j \in Y$, $p(y_i=y_j)$ is simply derived as the indicator function, which is 1 if equal, otherwise 0. We obtain a permutation-invariant loss by applying binary cross-entropy here:
\begin{align*}
    L(X, Y) &= \sum^N_i \sum^N_j [-p(y_i=y_j)\log{\sum^G_k p(g(x_i)=k)p(g(x_j)=k)} \\
            &- (1-p(y_i=y_j))\log{(1-\sum^G_k p(g(x_i)=k)p(g(x_j)=k))} \\
            &+ \alpha \sum^G_k p(g(x_i)=k)p(g(x_j)=k)]
\end{align*}
where $N=|X|=|Y|$. It can be efficiently computed by matrix multiplication as in Figure \ref{fig:architecture}. We introduce the last term to regularize group clumping, and $\alpha$ is its weight, set to $0.1$ in our experiment.

% 추가 인사이트
We sample $k \sim p(g(x_i))$ instead of $\text{argmax}_k\ p(g(x_i)=k)$ to get random outputs from the same input, which is often desired in prefracture method. $p(y_i=y_j)$ is the adjacency matrix of graph where vertices are $X$ and edges are group identity relations. It consists of several clique components representing each group. Our network can be interpreted as a probabilistic model of this graph. 

{
\captionsetup[figure]{skip=8pt}
\begin{figure}[t]
    \includegraphics[width=\linewidth]{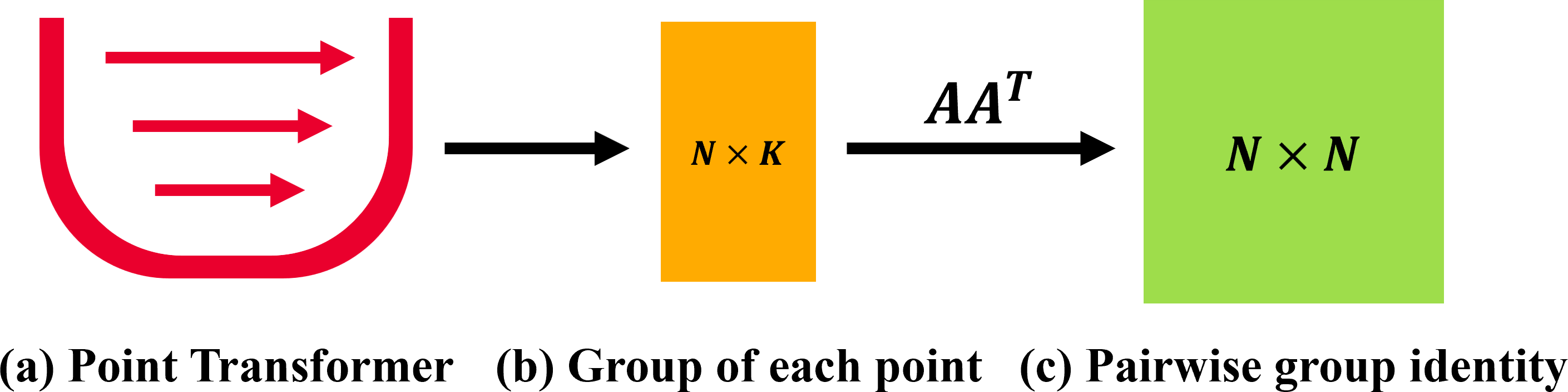}
    \caption{Proposed network architecture. (a) Point Transformer \cite{zhao2021point} as a backbone network. (b) The softmax of output as the probability that each point belongs to each group, where $N=|X|$ and $K=|G|$. (c) Matrix multiplication of (b) derives point-to-point group identity, providing permutation-invariant loss to group.
    }
    \Description{pictures of (a) U-net like architecture which denotes Point Transformer, (b) N by K matrix indicating the group of each point, and (c) N by N matrix that means pairwise group identity.}
    \label{fig:architecture}
\end{figure}
}

\section{Experimental Results}

% 데이터셋 선택과 사용
As our method does not have any geometric prior itself, the quality of dataset is important, still there are few datasets on object destruction. We train the network with Breaking Bad \cite{sellan2022breakingbad}, which is a synthetic dataset generated by physics-based simulation \cite{sellan2023breakinggood}. While it was originally proposed for 3D shape assembly, we flip its in/output pairs to supervise the network realistic object destruction. We break input mesh into fine pieces, and label them depending on the output fragment mesh containing their center of mass. The number of fragments is concatenated to the input, allowing control over the number of groups at inference.

% 결과 메쉬 뽑는 과정
To enable practical application, we post-process the group labels to obtain the prefractured mesh. First, we merges the pieces that are in a group and share a face into a single mesh. Second, we split non-adjacent pieces within a group into separate groups. The resulting group mesh set is ready-to-use, e.g., in UE 5.

% 결과가 좋다
The evaluation was conducted on unseen data from Objaverse \cite{deitke2023objaverse}. Figure \ref{fig:teaser} presents comparisons (a, b, c) and the showcase (d). Our approach outperforms existing methods, particularly in detecting structural weaknesses.

\section{Discussion}

We propose the novel paradigm exploiting the deep neural network for real-time object destruction, identifying structural weaknesses. However, limitations include the scarcity of datasets, and inadequacy of traditional segmentation benchmarks for evaluation. With emerging datasets like Fantastic Breaks \cite{lamb2023fantastic}, which provides material and real-world data but lacks in scale, future research building on ours may achieve more realistic destruction.

\begin{acks}
This work was supported by the IITP (Institute of Information \& Communications Technology Planning \& Evaluation) grant (National Program for Excellence in SW, 2023-0-00042 in 2024) funded by the Korea government (Ministry of Science and ICT).
\end{acks}

\bibliographystyle{ACM-Reference-Format}
\bibliography{bib}

\end{document}